# Memristor-Based Spiking Neural Network Accelerator for Bio-inspired Interception Task

Qianhou Qu
*Department of Electrical Engineering*
*The University of Texas at Arlington*
Arlington, USA
qxq9423@mavs.uta.edu

Sheng Lu
*Department of Electrical Engineering*
*The University of Texas at Arlington*
Arlington, USA
sxl2408@mavs.uta.edu

Liuting Shang
*Department of Electrical Engineering*
*The University of Texas at Arlington*
Arlington, USA
liuting.shang@mavs.uta.edu

Jaihan Utailawon
*Department of Electrical Engineering*
*The University of Texas at Arlington*
Arlington, USA
jxu9104@mavs.uta.edu

Sungyong Jung
*Department of Electrical Engineering and Computer Science*
*South Dakota State University*
Brookings, USA
Sungyong.Jung@sdstate.edu

Qilian Liang
*Department of Electrical Engineering*
*The University of Texas at Arlington*
Arlington, USA
liang@uta.edu

Chenyun Pan
*Department of Electrical Engineering*
*The University of Texas at Arlington*
Arlington, USA
chenyun.pan@uta.edu

***Abstract*— Spiking neural networks (SNNs) provide event-driven and low-power computation inspired by biological neural systems, but current implementations rely on von Neumann graphics processing units (GPUs) and central processing units (CPUs) platforms, where memory and computation bottlenecks limit energy efficiency. To address this challenge, this paper proposes an analog memristor-based spiking neural network (SNN) accelerator that integrates in-memory synaptic computation with analog integrate-and-fire (IF) neurons, eliminating multi-transistor CMOS synapse circuits and enabling asynchronous event-driven operation at the 45nm technology node. Additionally, a digital SNN accelerator is designed and optimized at the 5 nm technology node for comparison. The proposed architecture is evaluated using a predator–prey tracking task that emulates pursuit behavior. In this task, the analog SNN accelerator's inference closely matches the ideal software inference with a mean squared error (MSE) of 0.004. HSPICE simulation results show that the proposed analog SNN accelerator achieves 12.7× lower energy consumption and 1.26× lower delay compared to the digital baseline, demonstrating the potential of memristor-based neuromorphic circuits for energy-efficient real-time edge intelligence.***

## I. Introduction

The brain is widely regarded as an extremely energy-efficient information processing system [1]. It processes information through sparse, event-driven neural spikes exchanged between interconnected neurons and synapses. Spiking neural networks (SNNs), inspired by such a system, encode information through discrete spikes, enabling low-power and low-latency inference [2, 3]. However, unlike biological neural systems where synaptic weights are stored locally and computation is triggered only upon spike events, most spiking neural network tasks are executed on graphics processing units (GPUs) and central processing units (CPUs) following the von Neumann architecture for inference and training. The dense multiply-and-accumulate operations and frequent memory accesses [4, 5] lead to memory and compute bottlenecks that constrain the deployment of SNNs in edge and real-time systems. Therefore, designing hardware architectures that process information in a brain-like, event-driven manner is crucial for reducing energy consumption and improving efficiency.

Recently, numerous research efforts have focused on developing neuromorphic computing accelerators to support event-driven processing and spike-based computation. To enable energy-efficient event-driven inference, digital neuromorphic accelerators, such as TrueNorth [6] and Loihi [7] achieve scalable and energy-efficient spike-based computation. However, these digital neuromorphic accelerators still separate synaptic weight storage from computation and rely on clock-driven operation, rather than the asynchronous, spike-driven dynamics used by biological neurons.

In addition to digital neuromorphic accelerators, analog circuits and emerging devices have been explored for implementing brain-inspired neural processing [8-14]. In [9, 10, 12, 14], these approaches can leverage continuous-time dynamics and device physics to realize membrane integration, thresholding, and spike generation in an asynchronous and energy-efficient manner. However, during the membrane integration process, the accumulated membrane state can be influenced by the membrane capacitor directly connected to the input node, or by the internal state of emerging devices. In [11,

This manuscript is an author-prepared preprint version of a paper accepted for presentation at the 19th IEEE Dallas Circuits and Systems Conference.

13], the synaptic circuits are current-driven and decoupled from the postsynaptic membrane node, preventing membrane dynamics from influencing synaptic operation. However, the synapse implementation usually requires several transistors in circuits, which increases area overhead for synaptic arrays.

In this paper, we utilize memristors as synaptic devices for spiking neural network (SNN). Due to their nonvolatile and multi-level conductance characteristics, memristors provide synaptic weight storage and compact in-memory computation, replacing multi-transistor synapses. In addition, we design an integrate-and-fire (IF) neuron circuit that avoids the influence of the postsynaptic membrane voltage on spike generation, enabling clean event-driven operation. To demonstrate real-time neuromorphic inference, we evaluate the proposed architecture on a predator-prey tracking task.

The major contributions of this paper are listed below.

- We design and train a predator-prey tracking task to evaluate real-time neuromorphic inference performance.
- We design a memristor-based SNN circuit that integrates in-memory synaptic computation with an IF neuron for clean event-driven operation.
- We perform a comprehensive performance analysis and comparison by designing a digital SNN accelerator as a baseline.

## II. Network Overview

This section first introduces the predator-prey interception task and its sensory variables, then presents the SNN formulation and spike-based encoding scheme, and finally describes the supervised training methodology.

### A. Predator-Prey Tracking Task

Inspired by the interception model studied in [15], we design a two-dimensional predator-prey task that emulates the pursuit behavior of mammals. The predator and the prey move in a 2-D continuous space with time step-based motion and real-time control. At the beginning of each episode, the predator is initialized at the origin (0,0) while the prey is positioned at (10,0), resulting in an initial separation of 10 distance units. At every time step, the prey escapes with a constant speed of 1 unit/time step following a stochastic trajectory, while the predator pursues at a slightly higher speed of 1.2 units/time step. An interception task is considered successful when the predator closes the distance to 0.5 units or below.

To determine the turning angle at each step, the predator infers the prey's motion from four relative kinematic variables: previous distance $S_1$, current distance $S_2$, previous relative angle $\theta_1$ and current relative angle $\theta_2$, as shown in Fig. 1. These four variables serve as inputs to a 4-30-1 SNN that outputs a continuous steering angle for the next step, as shown in Fig. 2.

### B. Input and Output Encoding Scheme for SNN

To convert continuous kinematic variables into spike-based inputs suitable for SNN processing, we employ a deterministic rate-coding scheme. Each of the four inputs is first normalized into the range [0,1] and then mapped into a discrete spike train over a fixed simulation window. The maximum number of spikes allowed for any input neuron is set to 25, and the numerical value is encoded through the corresponding spike count. The network output follows the same rate-coding principle.

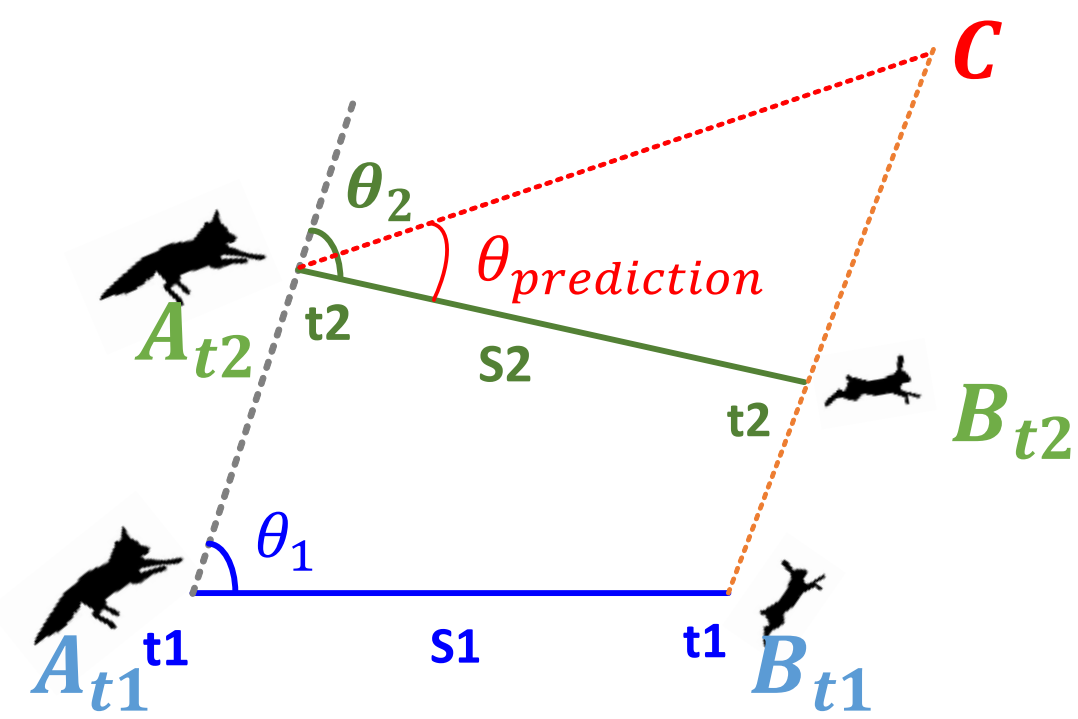

Fig. 1 SNN task representation showing relative distance and bearing angles used to predict the next turning angle.

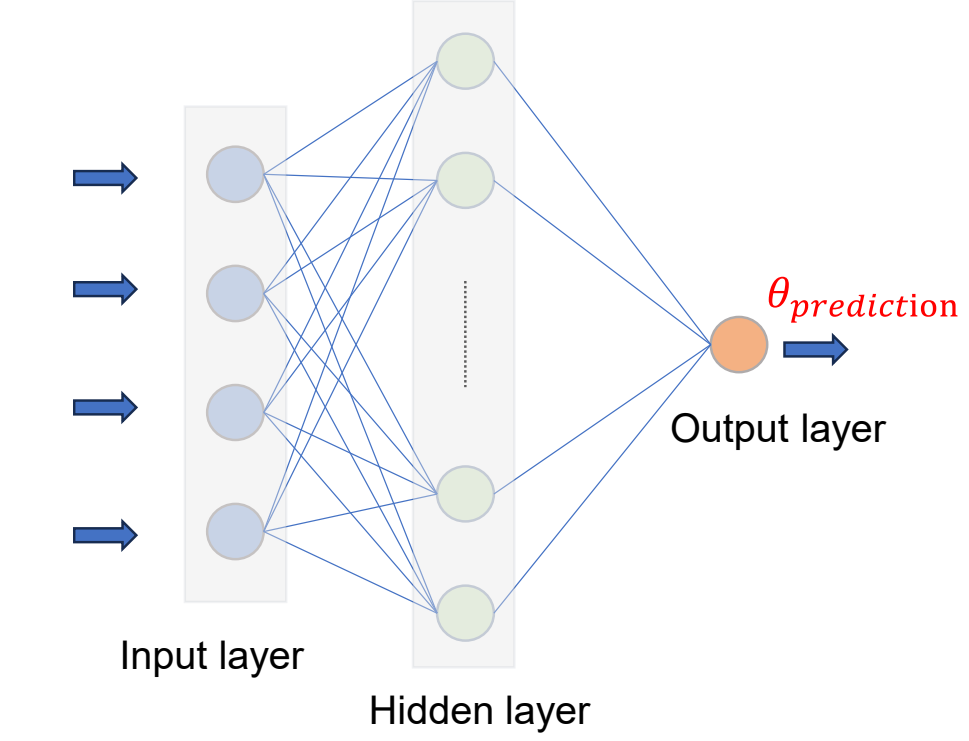

Fig. 2 SNN architecture for predator pursuit task.

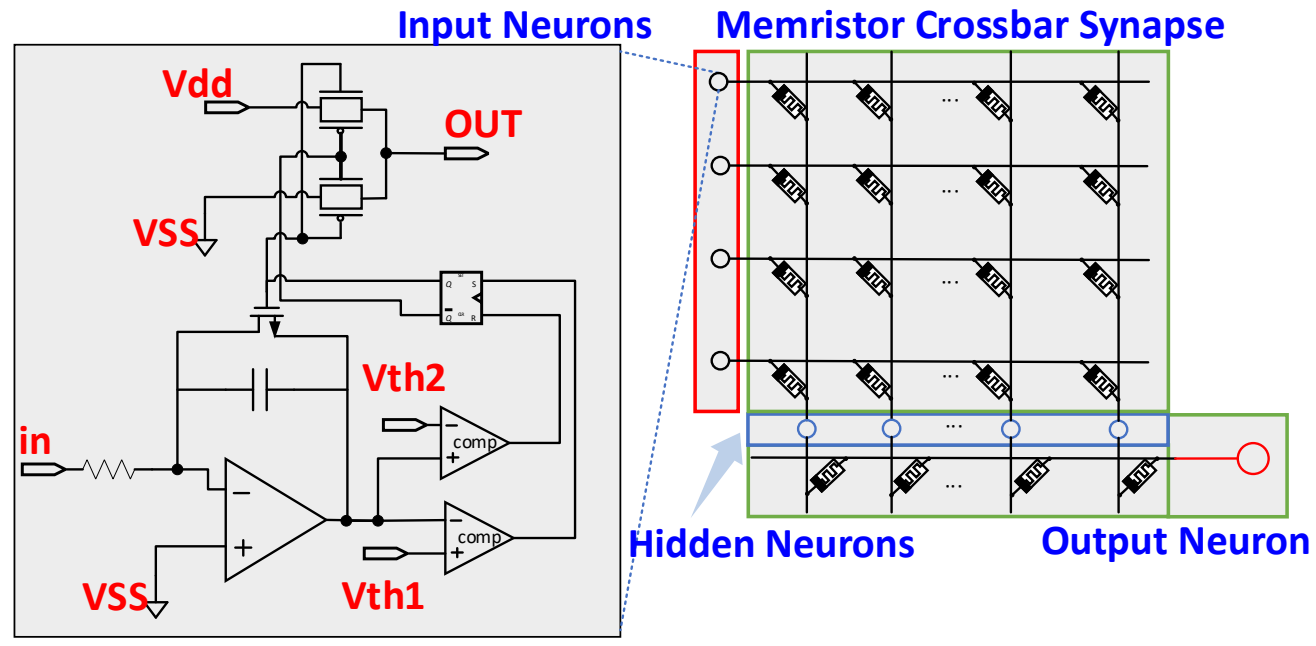

Fig. 3 Analog SNN accelerator circuit.

### C. Training Samples and Methodology

Ideal training labels are generated using a geometrical interception model. Let $A_t$ and $B_t$ denote the positions of the predator and prey at time t. For each time step, an interception point C is defined such that the arrival times of the predator and prey at C are identical. This interception condition ensures that the predator does not perform pure pursuit of the prey's instantaneous position but instead leads the prey toward a future collision point C. Therefore, the angle $\angle CAB$ formed by the interception point C, the predator position $A_t$, and the prey position $B_t$ represents the ideal steering angle that the predator should take at time t. The SNN is trained in a supervised manner using SNNTorch [16] with gradient backpropagation through time (BPTT) and an MSE loss between the predicted and ideal turning angles.

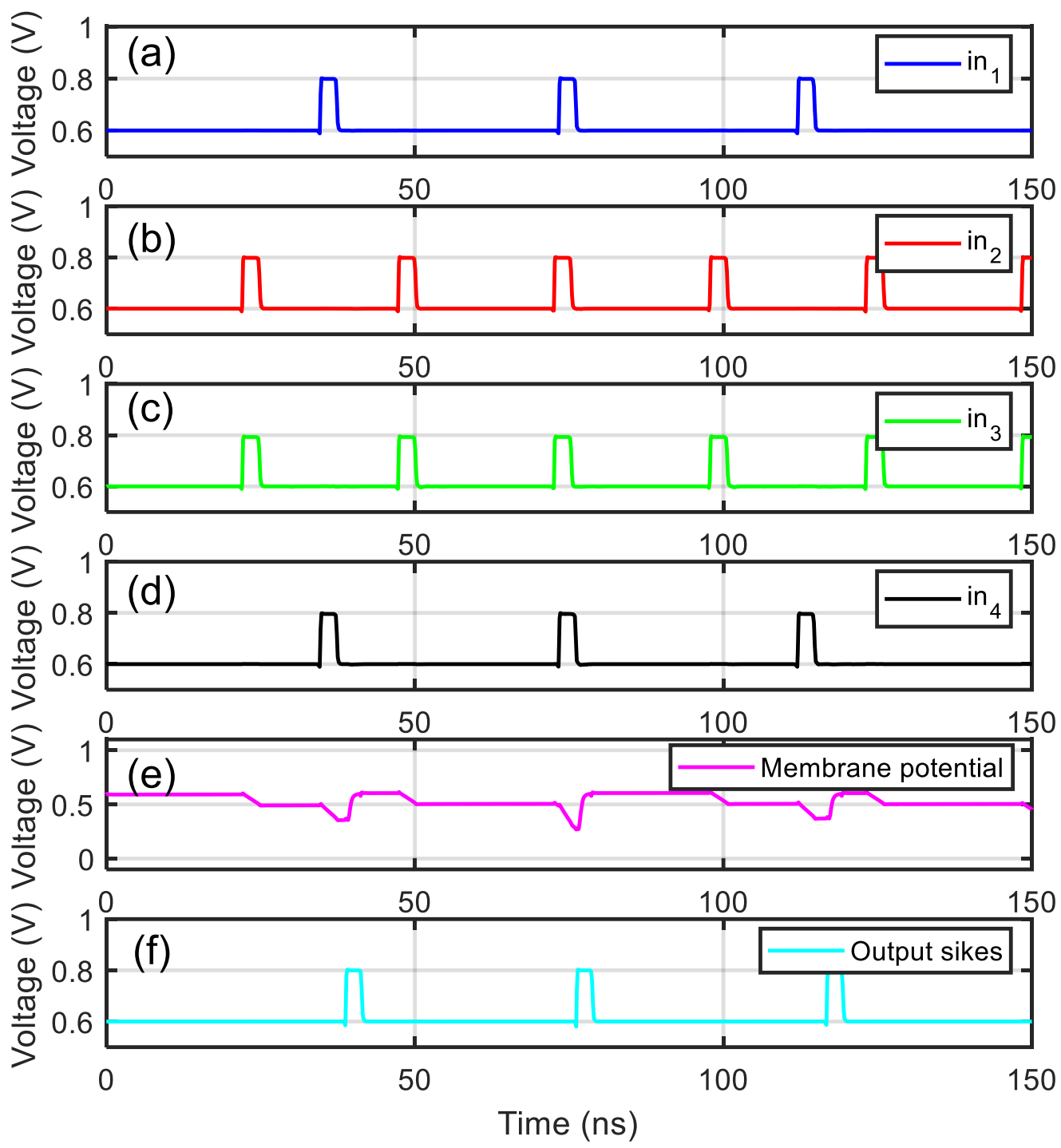

Fig. 5 Transient waveforms of (a) – (d) spikes at four inputs, (e) membrane potential, and (f) output neuron spike of a hidden neuron.

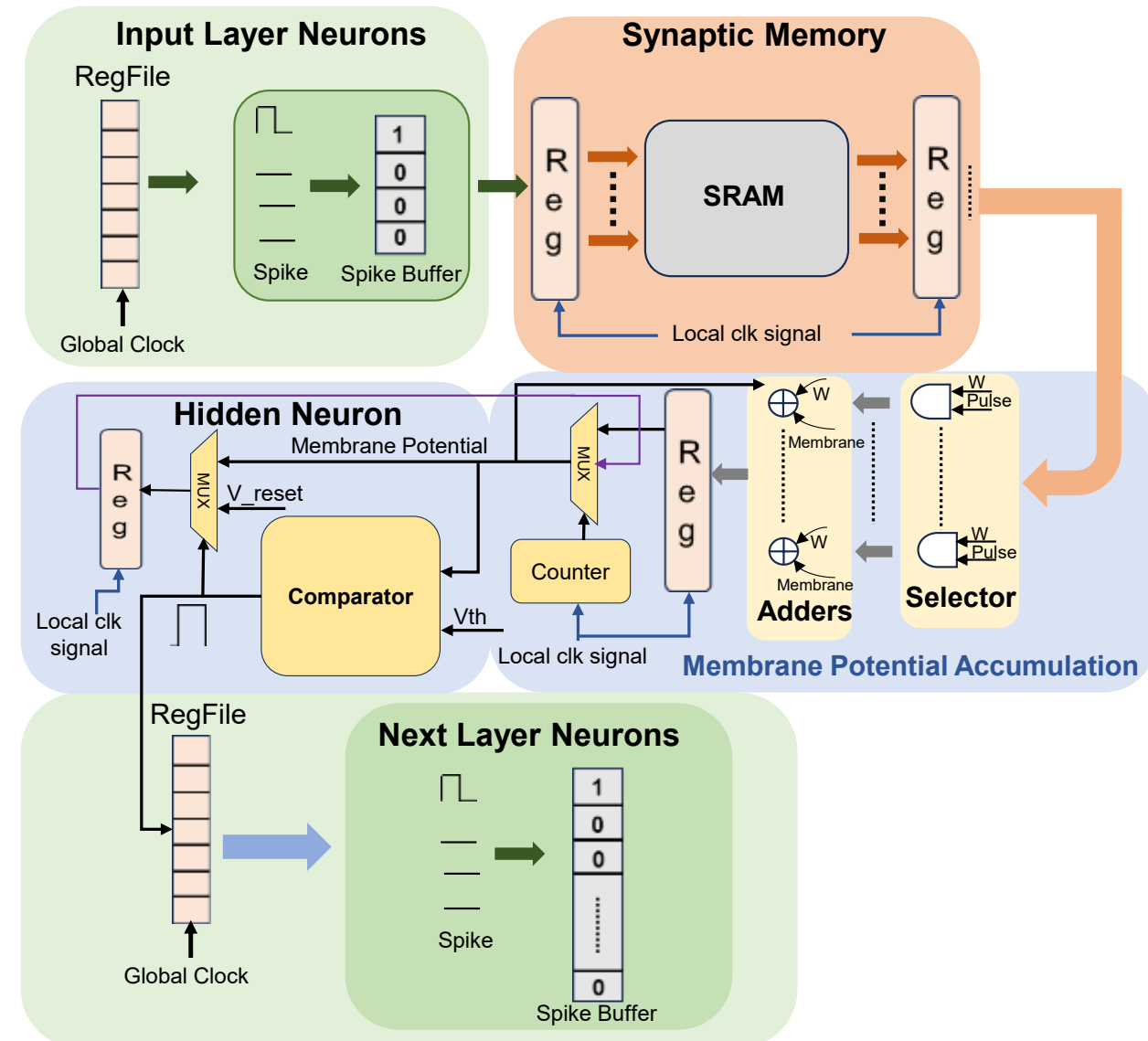

Fig. 4 Digital SNN accelerator architecture.

## III. Proposed Design and Modeling Approach

Based on the SNN model presented in Section II, this section first introduces the proposed analog accelerator for SNN inference, followed by a digital accelerator that serves as a reference design for comparison.

### A. Analog Circuit-based SNN Accelerator

Fig. 3 shows the proposed architecture of the analog SNN accelerator, consisting of 4 input IF neurons, 30 hidden neurons, and 1 output IF neuron. Synaptic connections between layers are implemented using memristor crossbar arrays, which store synaptic weights and perform in-memory computation for membrane voltage accumulation.

The neuron circuits (Fig. 3, left) integrate the synaptic input currents through an op-amp-based integrator that generates the membrane potential. Due to the virtual-ground behavior at the op-amp input node, the membrane potential does not interfere with the input summation, enabling stable current integration across incoming spike trains without attenuation. The resting membrane potential is set to 0.6V, and the firing threshold $V_{th1}$ is set to 0.4V. Two comparators are used to detect threshold crossing and enforce membrane reset. The first comparator monitors whether the membrane potential exceeds the firing threshold and triggers the reset phase. However, using only a single comparator would disable the reset path once the membrane potential settles above the threshold due to downward discharging. To ensure a full reset, a second comparator detects whether the membrane voltage has reached the desired reset level $V_{th2}$. A latch then maintains the reset control signal, allowing the NMOS reset transistors connected across the integrator node to fully discharge the membrane. Finally, two transmission gates, controlled by the latch, generate the output spike waveform at node OUT.

As shown in Fig. 3, the memristor crossbar synapses operate in a conductance range from $G_{min} = 0.2\ \mu S$ to $G_{max} = 10\mu S$, corresponding to a conductance ratio of 50 and supporting 100 discrete conductance levels across the range. To map conductance to synaptic weights, a presynaptic spike is injected for each conductance level and the resulting membrane potential increment at the postsynaptic neuron is measured. The effective synaptic weight for conductance level $G_i$ is then given by

$$w_i = \frac{\Delta V_{mem}(G_i)}{V_{rest} - V_{th1}}, \quad (1)$$

where $\Delta V_{mem}(G_i)$ denotes the membrane voltage swing induced by $G_i$, $V_{rest}$ is the resting potential when the neuron is silent and $V_{rest} - V_{th1}$ represents the maximum membrane integration window.

Fig. 5 shows the time-domain behavior of the proposed analog integrate-and-fire neuron obtained from HSPICE simulations. Four presynaptic spike trains are injected into the neuron, each generating a current pulse through the crossbar synapse. As the spikes arrive, the membrane voltage integrates downward until the firing threshold is reached, at which point the reset mechanism discharges the membrane and generates an output spike. The waveform confirms the expected integrate-and-fire dynamics, including integration, threshold detection, reset, and spike generation.

### B. *Digital Circuit-based Accelerator*

Fig. 4 illustrates the digital circuit implementation that is used as a baseline for comparison. Input spikes are stored in a register file and buffered across time steps. Synaptic weights are stored in a static random-access memory (SRAM) based synaptic memory and are represented using a 16-bit unsigned fixed-point format, consisting of 1 integer bit and 15 fractional bits to support fine-grained weight precision. A selector performs a logical AND operation between the spike event at the current time step and the stored synaptic weight, effectively computing the spike-weight product required for SNN inference. The circuit operates in a clocked pipeline. A global clock synchronizes spike buffering, threshold detection, and inter-layer data movement, while a local clock sequentially scans the synaptic inputs and accumulates their corresponding weights into the membrane potential. This time-multiplexed accumulation removes the need for parallel accumulation and reduces memory bandwidth and area overhead.

For each hidden neuron, a digital comparator checks whether the membrane potential exceeds a predefined threshold to generate a spike event. After a spike is generated, the membrane potential is reset to $V_{rest}$, enabling the next integration cycle. The output spike count is tracked by a counter, constituting the rate-coded representation of the neuron output. Spikes are then latched and forwarded to the next layer through the same buffering mechanism.

## IV. Simulation Results

### A. *Application-Level Demonstration and Accuracy Analysis*

To demonstrate functional fidelity, we evaluate whether the analog accelerator preserves SNN inference accuracy at both step-level prediction and trajectory-level behavior. Results are compared against a 16-bit digital accelerator and SNNTorch baseline.

Fig. 6 shows the prediction accuracy of both the digital and analog SNN accelerators against the software SNN baseline. Each scatter point corresponds to one time-step prediction, yielding a total of 1300 evaluation samples. The diagonal y = x line denotes the ideal output, and prediction accuracy increases as points approach this line. The digital accelerator performs inference using 16-bit fixed-point arithmetic with no floating-point computation, while the proposed analog accelerator results are obtained from full-circuit HSPICE simulations. The analog accelerator exhibits strong consistency with the software baseline, demonstrating high correlation (R = 0.98) and low mean-square error (MSE = 0.004), indicating that the compute-in-memory and analog neuron circuits introduce minimal deviation in task inference accuracy.

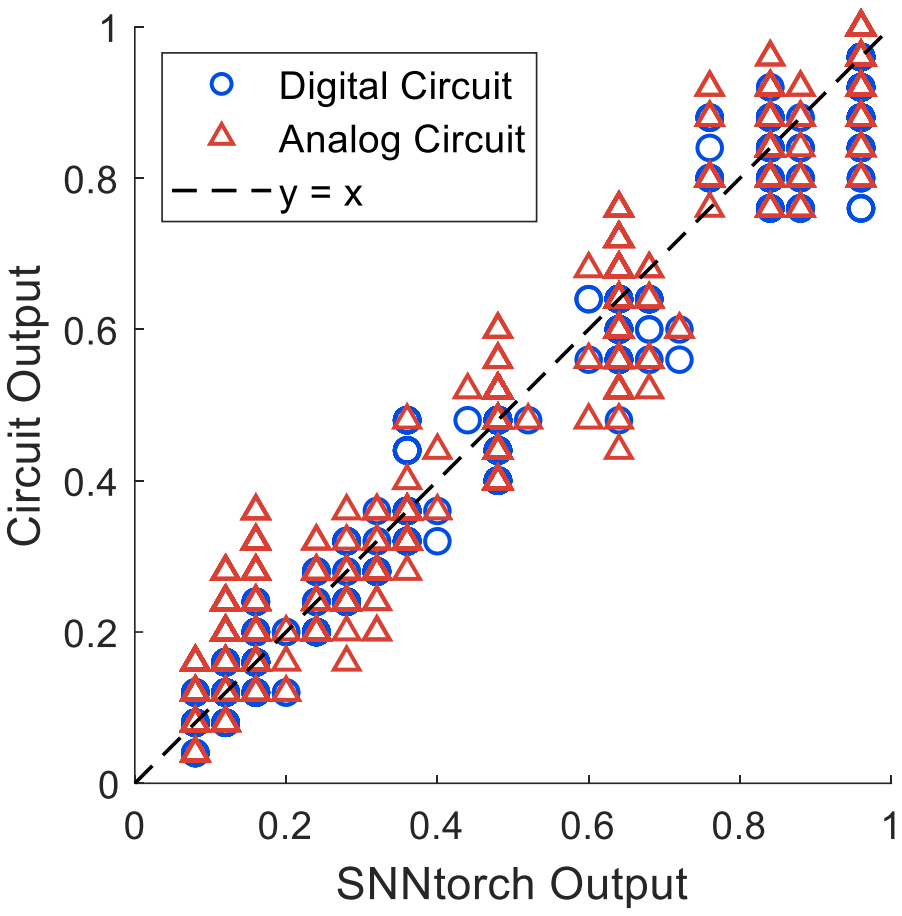

Fig. 6 Comparison of digital and analog accelerator outputs against SNNTorch results.

Fig. 7 shows two representative predator-prey interception scenarios. The prey trajectory is plotted together with the inferred trajectories generated by the analog and digital accelerators. In both examples, the predator successfully intercepts the prey, and the analog trajectory closely overlaps with the digital trajectory, demonstrating consistent decision-making dynamics. These results confirm that the proposed analog accelerator preserves task-level performance comparable to the digital accelerator while operating under device-level circuit constraints.

### B. *Energy Comparison*

To fairly evaluate the energy efficiency of the digital and proposed accelerators, we define the per-spike energy for both accelerators by activating one input neuron while keeping the remaining inputs silent. This metric captures the synaptic accumulation and membrane update of all 30 hidden neurons for one inference step.

For the proposed analog accelerator, the energy is obtained from HSPICE, which is dominated by the integrator (36%) and the comparators (62%). For the digital accelerator, the energy is estimated using IRDS [17] 5nm FinFET device parameters. Standard-cell energy is first derived from the transistor-level switching characteristics, and then the energy of each functional block can be estimated based on the cell usage. The dominant digital energy comes from (i) register-file switching activity between pipeline stages during each clock cycle and (ii) fixed-point adder for the membrane accumulation during each spike activity. Fig. 8 shows that the proposed analog SNN accelerator achieves 12.7× lower energy consumption. The improvement mainly comes from current-mode in-memory accumulation and continuous-time analog neuron dynamics, eliminating multi-bit accumulation and wide memory access.

### C. *Delay Comparison*

We evaluate latency using the same activation condition as in the energy analysis: a single active input spike while all remaining inputs remain silent. The delay corresponds to the time required for the 30 hidden neurons to complete membrane potential accumulation for one inference step.

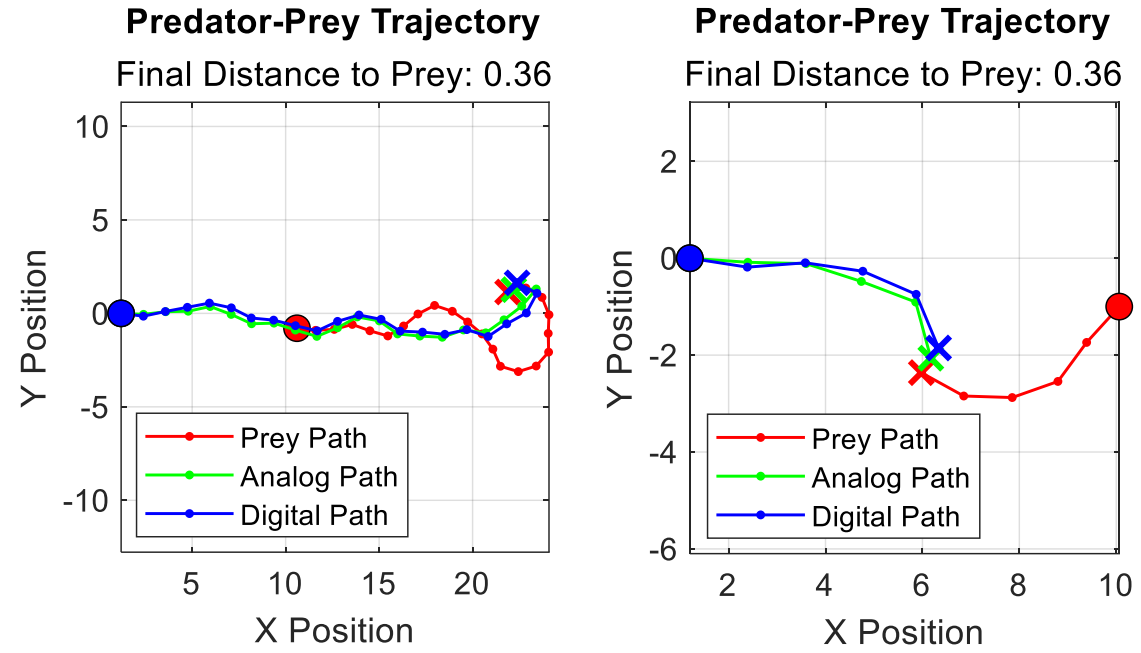

Fig. 7 Trajectory comparison between analog and digital accelerator.

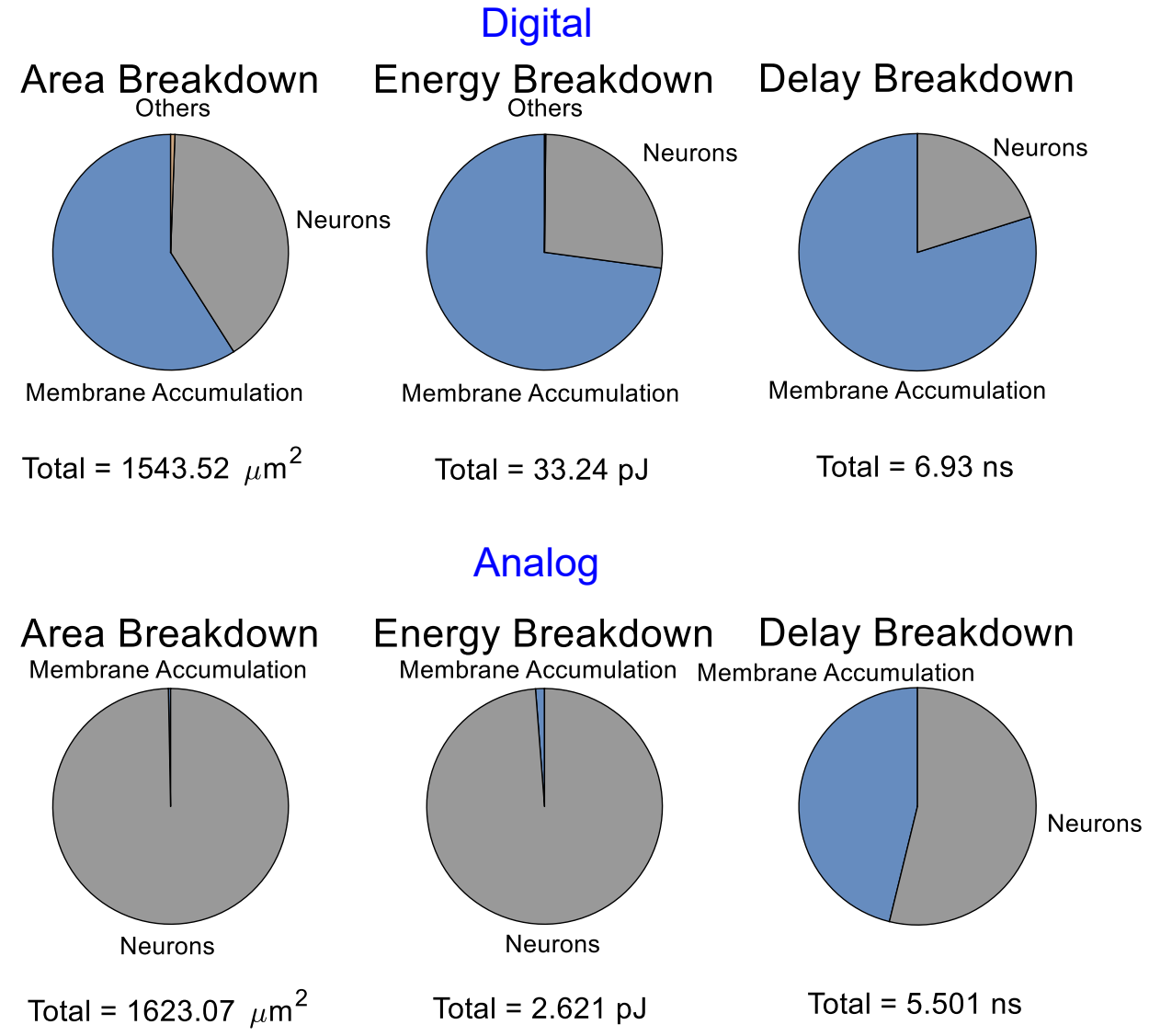

Fig. 8 Area, energy, and delay comparison between digital and analog SNN accelerators.

For the analog accelerator, delay is extracted from HSPICE by measuring the worst-case propagation path through the memristor crossbar and the neuron circuit. The delay consists of membrane accumulation followed by comparator-triggered neuron dynamics, contributing roughly comparable fractions. For the digital accelerator, delay is estimated using IRDS [17] 5nm FinFET device parameters. Transistor switching delays are translated into standard-cell delays, and the total path delay is obtained by aggregating the cell delays within each pipeline stage. Since computation is pipelined, the critical latency is determined by the longest stage. The SRAM read path dominates the digital delay. As shown in Fig. 8, the proposed analog SNN accelerator achieves 1.26× lower delay compared to the digital baseline. The lower analog latency is attributed to continuous-time accumulation and the absence of SRAM fetch and multi-bit pipeline stages.

### D. Area Comparison

Area is evaluated by estimating the silicon footprint of the accelerator blocks. For the analog accelerator, we use the NCSU FreePDK 45nm [18] design rules. A NAND2 gate is used as the unit reference cell, and the device areas of the op-amp, comparators, and reset circuitry are approximated by scaling the transistor widths relative to the NAND2 reference. Interconnect area is estimated from pin counts using the metal-pitch and spacing rules. The memristor crossbar area is computed assuming BEOL stacking between M1 and M2 layers, and follows the FreePDK 45nm metal rules.

For the digital accelerator, standard-cell footprints are first extracted from the ASAP7 FinFET PDK [19] and then scaled to a 5nm FinFET technology node. Functional block area (SRAM, adders, registers, and control logic) is obtained by summing the required cells per block. The resulting area breakdowns are shown in Fig. 8. At the 45nm technology node, the analog accelerator occupies slightly larger area compared to the 5nm scaled digital implementation, mainly due to the analog neuron circuits, such as opamp and comparator.

## V. Conclusion

In this work, we propose an analog SNN accelerator that combines memristor compute-in-memory with analog neuron circuits. The accelerator is evaluated on a predator-prey task and demonstrates high consistency with the software SNN baseline, preserving trajectory-level interception behavior. Hardware evaluation shows that the proposed analog accelerator achieves significantly lower energy and delay compared to a 16-bit digital baseline, while maintaining comparable area. These results indicate that analog neuromorphic computing provides an efficient and low-latency solution for real-time inference tasks.